\documentclass[10pt, a4paper]{article}

\usepackage{lrec-coling2024} 

\title{RAAMove: A Corpus for Analyzing Moves in Research Article Abstracts\\ \vspace*{.5\baselineskip}}

\name{Hongzheng Li$^{\ast}$ Ruojin Wang$^{\ast}$\thanks{{$\triangle$}The first two authors contributed equally to this work.}, Ge Shi$^{\dagger}$, Xing Lv$^{\ast}$, Lei Lei$^{\ast}$, Chong Feng$^{\ddagger}$, \\ {\bf \large Fang Liu$^{\ast}$, Jinkun Lin$^{\ddagger}$, Yangguang Mei$^{\dagger}$, Lingnan Xu$^{\ddagger}$}}

\address{$^{\ast}$School of Foreign Languages,\\
        Key Laboratory of Language, Cognition and Computation Ministry of Industry and Information Technology,\\ 
        Beijing Institute of Technology\\
        $^{\dagger}$Faculty of Information Technology, Beijing University of Technology \\
        $^{\ddagger}$School of Computer Science, Beijing Institute of Technology \\
        $^{\ast}$8, Liangxiang East Road, Fangshan District, 102488, Beijing \\
        $^{\ddagger}$5, Zhongguancun South St., Haidian District, 100081, Beijing, China\\
        $^{\dagger}$100, Pingleyuan, Chaoyang District, 100124, Beijing, China\\
          \{lihongzheng,wangruojin,lvxing,leilei,fengchong,liufang,ljkun,xln\}@bit.edu.cn\\
           \{shige,meiyangguang\}@bjut.edu.cn}

\abstract{
Move structures have been studied in English for Specific Purposes (ESP) and English for Academic Purposes (EAP) for decades. However, there are few move annotation corpora for Research Article (RA) abstracts. In this paper, we introduce RAAMove, a comprehensive multi-domain corpus dedicated to the annotation of move structures in RA abstracts. The primary objective of RAAMove is to facilitate move analysis and automatic move identification. This paper provides a thorough discussion of the corpus construction process, including the scheme, data collection, annotation guidelines, and annotation procedures. The corpus is constructed through two stages: initially, expert annotators manually annotate high-quality data; subsequently, based on the human-annotated data, a BERT-based model is employed for automatic annotation with the help of experts' modification. The result is a large-scale and high-quality corpus comprising 33,988 annotated instances. We also conduct preliminary move identification experiments using the BERT-based model to verify the effectiveness of the proposed corpus and model. The annotated corpus is available for academic research purposes and can serve as essential resources for move analysis, English language teaching and writing, as well as move/discourse-related tasks in Natural Language Processing (NLP).
 \\ \newline  }

\begin{document}

\maketitleabstract

\section{Introduction}
Effective communication in research articles relies on the use of "\textit{moves}", which are discoursal semantic units serving specific functions in written or spoken discourse, and linked to the writer’s purpose \cite{Swales:1990,Swales:2004}. A move represents text frames that have their own meaning and function \cite{Connor-Upton-Kanoksilpatham:2007}.

In the realm of scientific research, Research Articles (RAs) play an important role in fostering academic exchange and promoting scientific progress.
Within the RA, the abstract serves as an indispensable section, conveying the core ideas and highlights of the entire article in a concise yet convincing manner.
Good abstracts should include well-organized move structures to promote the importance and values of the article. Just as \citet{Hyland:2007} argued, "One way that writers claimed significance was by opening their abstracts with a promotional statement."

Here is an abstract example\footnote{The abstract example is from \href{https://aclanthology.org/2022.acl-long.65/}{Su et al., ACL 2022}} from the field of Natural Language Processing (NLP). The abstract contains clear and basic moves in square brackets, making the abstract highly readable and effectively showing the author's ideas and contributions.

\setlength{\parindent}{20pt}

[\textbf{Background}]Large-scale pretrained language models have achieved SOTA results on NLP tasks. [\textbf{Gap}]However, they have been shown vulnerable to adversarial attacks especially for logographic languages like Chinese. [\textbf{Purpose}]In this work, we propose RoCBert: a pretrained Chinese Bert that is robust to various forms of adversarial attacks like word perturbation, synonyms, typos, etc. [\textbf{Method}]It is pretrained with the contrastive learning objective which maximizes the label consistency under different synthesized adversarial examples. ...... [\textbf{Result}]Across 5 Chinese NLU tasks, RoCBert outperforms strong baselines under three blackbox adversarial algorithms without sacrificing the performance on clean testset. 

\setlength{\parindent}{0pt}


Move analysis has great potential for instructing, assessing abstract writing, aiding English for Specific Purposes (ESP) and English for Academic Purposes (EAP) learners and writers \cite{
hyland2008can,moreno2018strengthening,swales2019futures}, which has attracted global attention for decades, with numerous studies investigating moves, grammatical and linguistic features in sections of RAs such as Introduction and Method \cite{cotos2017move,lu2021matching,alsharif2023rhetorical}. Some previous studies have also focused on moves and linguistic forms in abstracts \cite{Pho:2008,suntara2013research,darabad2016move}, and some work on move detection and recognition \cite{dayrell-etal-2012-rhetorical,ding2019factors}.

However, previous studies have predominantly focused on the analysis of moves within specific domains or journals,  with few addressing multi-disciplinary RAs. For example, some works have discussed the rhetorical moves and there variations in Applied Linguistics journals \cite{Pho:2008,SamarTalebzadehKianyAkbari:2014,yoon2020rhetorical}. To the best of our knowledge, there is a limited amount of research on the automated annotation of discourse moves in academic writing, and there is also a scarcity of large-scale annotated corpora specifically designed for move annotation within Research Article (RA) abstracts. This insufficiency poses a barrier to meeting the growing demand for comparative research in move analysis across diverse domains and tasks in Natural Language Processing (NLP), including the automatic identification of discourse moves and Computer-Assisted Language Learning (CALL).


To address this research gap, our long-term goal, which centers on intelligent English writing assistance and text generation for intelligent and digital language education, supports our mid-term goal of automatic move structure identification. In this pursuit, we introduce the RAAMove\footnote{The RAAMove corpus is available at \href{https://github.com/ljk1228/RAAMove}{https://github.com/ljk1228/RAAMove}}, a large-scale multi-domain corpus of MOVE structures in the RA Abstracts from three scientific disciplines: Artificial Intelligence (AI), Communication Engineering (CE), and Mechanical Engineering (ME), where the latter two can be collectively referred to as Engineering. The corpus is constructed in two stages: In the first stage, the data is manually annotated to form a high-quality dataset; in the second stage, a simple but efficient BERT-based model is trained to achieve automatic annotation. Manual correction is then performed to modify the model's error annotation. We also conduct automatic move identification experiments to testify the built corpus. The results underscore the effectiveness of the annotation model and the value of the corpus.

The contributions of our paper are as follows:
\begin{itemize}
    \item We develop a multi-domain move structure annotation corpus for analyzing moves in RA abstracts. This can benefit non-English speakers and students, helping them understand move structures and improve their writing skills. In addition, it can also serve as an essential and valuable resource for language teaching and learning, as well as for other tasks and applications related to move and discourse analysis.
    \item We suggest a revision of move structure categories based on Hyland's established move classification \cite{Hyland_2000}, which is demonstrated to be better suited for the annotation of moves within scientific Research Article (RA) abstracts and engineering practice.
    \item We propose an innovative BERT-based automatic annotation model that incorporates word-level saliency attribution. The model is beneficial for facilitating the expansion of the corpus size while maintaining annotation quality.
\end{itemize}

The remaining of this paper is organized as follows: Section 2 discusses related work, Section 3 describes the scheme, and Section 4 presents the construction of the corpus in detail. Section 5 gives the corpus statistics. Section 6 conducts the experiments and analysis, and Section 7 offers a conclusion.

\section{Related Work}

\subsection{Scholarly Dataset and Corpora}
Corpora of academic texts contain scholarly writing such as research papers, essays and abstracts published in academic journals, conference proceedings, etc. Building scholarly corpora is beneficial for promoting scientific progress and have practical values in many application scenarios, attracting widespread attention from the academic community.

The number of scholarly datasets has increased significantly in recent years. \citet{Rohatgi-Qin-Aw-Unnithan-Kan:2023} argue that some of them are task-specific corpora which include selective information of scientific papers \cite{hsu-etal-2021-scicap-generating,hou-etal-2021-tdmsci,cachola-etal-2020-tldr}; some others are open research-aimed datasets. These include corpora built based on \textit{ACL Anthology}, such as ACL OCL corpus \cite{Rohatgi-Qin-Aw-Unnithan-Kan:2023}, S2ORC \cite{lo-etal-2020-s2orc}, ARC \cite{bird-etal-2008-acl} and AAN \cite{radev-etal-2009-acl}. These datasets usually contain comprehensive metadata and full text of scientific papers, providing researchers with accessible and essential data resources.

\subsection{Move Structure Annotation}
Analysis of arguments and rhetorical move structures in articles has a long history. The Create-A-Research-Space (CARS) model proposed by Swales’ pioneering work \cite{Swales:1990,Swales:2004} and move categories (an abstract can consist of up to five
possible moves)\cite{swales2009abstracts} had widespread influence and gave rise to a plethora of studies of academic discourses in terms of their move structures, but also had some problems. Other improved theoretical approaches like move analysis \cite{Kanoksilapatham:2003}, Argumentative Zoning (AZ) scheme \cite{teufel-etal-1999-annotation} and revised AZ versions \cite{teufel2009towards,teufel2010structure} propose useful guidelines for annotating moves in scientific articles.  

Recently some datasets have been built for studying the move structures. \citet{alliheedi-etal-2019-annotation} focuses on the semantic roles and rhetorical moves in the Methods section of biochemistry articles.  \citet{liu:2016}constructed a dataset of abstracts selected from the Applied Linguistics Journal; \citet{dayrell-etal-2012-rhetorical}constructed a corpus of abstracts in the fields of Physical Sciences and Engineering, Life and Health Sciences. \citet{cortes2013purpose}examined the use of lexical bundles and their correlation with move steps with a corpus of 1,300 published RA introductions. \citet{tovar2020rhetorical} investigated the rhetorical organization of RA abstracts published in native and non-native English-speaking countries by building an abstract corpus. 

Considering that the number of corpora specifically for the moves in abstracts is still small, we believe that the creation of a comprehensive, large-scale corpus is indeed essential. Therefore, the following sections will discuss the scheme and methodology behind our construction of an academic corpus, drawing inspiration from prior theoretical and practical achievements.

\section{Scheme}
This section attempts to answer the following questions:
\begin{itemize}
    \item Which rhetorical move theories could guide the construction of a corpus for analyzing moves in RA abstracts?
    \item In what manner can these theories find greater relevance in our work?
\end{itemize}
In light of this, this section begins by introducing Hyland's move classification \cite{Hyland_2000}, which serves as the theoretical foundation underlying our research, then proceeds with a pilot study, and finally, modifies this classification to meet the needs of constructing a corpus for analyzing moves in current RA abstracts. Unless otherwise stated, Hyland's move classification will be referred to as \textit{H-2000} below.

\subsection{Hyland’s Move Classification}

The rhetorical move structures appearing in RAs have been thoroughly analyzed in prior studies. \citet{Hyland_2000}, nonetheless, discerned that the move classification observed in full-length RAs might not be entirely suitable for analyzing RA abstracts. Consequently, he delved into a concentrated yet exhaustive investigation, with a particular focus on categorizing the move structures within RA abstracts. His analysis encompasses 1,400 texts across eight academic disciplines, which can be further categorized into pure sciences, applied sciences, humanities, and social sciences, deliberately covering a wide range of academic practices. This classification stands as a significant contribution for its critical insight into the intricacies and nuances of abstract writing practices and conventions in various disciplines, and thus, has been adopted as the theoretical foundation of this work.

An overview of H-2000 is presented in Table~\ref{tab:table1}.

\begin{table*}[htbp]
    \centering
    \begin{tabular}{|p{0.1\textwidth}|p{0.8\textwidth}|}
    \hline
        \textbf{Move} & \textbf{Function}\\
    \hline
        Introduction & Establishes context of the paper and motivates the research or discussion. \\
    \hline
        Purpose & Indicates purpose, thesis or hypothesis, outlines the intention behind the paper. \\
    \hline
        Method & Provides information on design, procedures, assumptions, approach, data, etc. \\
    \hline
        Product & States main findings or results, the argument, or what was accomplished.\\
    \hline
        Conclusion & Interprets or extends results beyond scope of paper, draws inferences, points to applications or wider implications.\\
    \hline
    \end{tabular}
    \caption{Hyland’s classification of rhetorical moves in RA abstracts}
    \label{tab:table1}
\end{table*}

\subsection{Pilot Study}

Conducting a pilot study, as emphasized by \citet{biber_connor_upton_2007}, is deemed essential in corpus-based genre analysis to ascertain the study’s feasibility. In this context, this pilot study was undertaken to assess the applicability of H-2000 for RA abstracts within the disciplines of NLP and Engineering. Furthermore, it sought to uncover any unforeseen issues that might hinder our exploration of these abstracts’ move structures and construction of the move corpus. 

Our pilot study examined a total of 40 abstracts, with 20 selected from each of the aforementioned disciplines. The data were collected from two highly influential top-tier publications: \textit{the International Journal of Heat and Mass Transfer}\footnote{https://www.sciencedirect.com/journal/international-journal-of-heat-and-mass-transfer}, as well as the \textit{Proceedings of the 59th ACL-IJCNLP (Volume 1: Long Papers)}\footnote{https://aclanthology.org/events/acl-2021/\#2021acl-long}. These abstracts were manually annotated with H-2000 move categories as illustrated in Table~\ref{tab:table2}.

\begin{table}[htbp]
    \centering    \begin{tabular}{|c|c|c|c|c|}
    \hline
 & \multicolumn{2}{c|}{\textbf{AI}}& \multicolumn{2}{c|}{\textbf{Engineering}}\\
    \hline
        \textbf{Move} & \textbf{Freq.}& \textbf{\%}& \textbf{Freq.}&\textbf{\%}\\
    \hline
        Intro.& 17& 85\% & 16&80\% \\ 
        Pur.& 20& 100\% & 19&95\% \\ 
        Met.& 19& 95\% & 19&95\% \\ 
        Pro.& 11& 55\% & 16&80\% \\ 
        Con.& 11& 55\% & 13&65\% \\
    \hline
    \end{tabular}
    \caption{Frequency of moves identified based on Hyland’s classification in sample abstracts. Where the five moves correspond to those in Table~\ref{tab:table1}.}
    \label{tab:table2}
\end{table}

H-2000 closely aligns with the moves identified in the abstracts we annotated, particularly within the Introduction, Purpose, and Method categories. Noticeable distinctions, however, become evident in the Product and Conclusion categories. His definition of Product encompasses not only the statement of findings and results but also the presentation of arguments. However, we did not encounter a single sentence that fulfilled the communicative function of presenting arguments during our pilot study. 
It is also worth noting that directly applying the Conclusion category to analyze the move structures of abstracts in these fields may appear inappropriate. Consider the following instances:

\setlength{\parindent}{20pt}
[\textbf{Example 1}]: These results can provide a better understanding of surfactants and guide the practical preparation of multicomponent fluids for boiling heat transfer enhancement.

\setlength{\parindent}{20pt}
[\textbf{Example 2}]: We release source code for our models and experiments at https://github.com/xxx.

\setlength{\parindent}{0pt}
Sentences resembling example 1 may not seamlessly fit within the Conclusion category, as this move typically emphasizes the interpretation and extension of the results rather than pointing out potential effects on related studies and practical applications. Sentences like example 2 often appear in the concluding sections of abstracts in the field of NLP, denoting a willingness to release research code or datasets to assist colleagues and foster future research. These sentences also fail to align with any of the H-2000 categories. 

Moreover, sentences falling under the Introduction category in H-2000 appear to serve diverse communicative purposes. Consider the examples 3 and 4, where the former introduces the task, and the letter outlines deficiencies in recent studies: 

\setlength{\parindent}{20pt}
[\textbf{Example 3}]: Undermining the impact of hateful content with informed and non-aggressive responses, called counter-narratives, has emerged as a possible solution for having healthier online communities. 

\setlength{\parindent}{20pt}
[\textbf{Example 4}]: Although such studies have made an effort to build hate speech / counter-narrative (HS/CN) datasets for neural generation, they fall short in reaching either high-quality and/or high-quantity.

\setlength{\parindent}{0pt}
The findings from this pilot study underscore the need for a more detailed move structures’ classification. It also reveals significant differences between the abstracts in these two disciplines, highlighting the importance of establishing domain-specific move corpora to better tailor future applications.

\subsection{Modifying the Move Categories}

Considering the primary objective of our research, which is the automated identification of rhetorical moves in RA abstracts, conducting a more precise categorization and annotation of moves can facilitate a more thorough examination of their organizational structure and underlying logic. In this context, our pilot study prompted four modifications to H-2000. The modified classification is outlined in Table~\ref{tab:table3}. To begin, we proposed two distinct categories to replace the Introduction category: Background and Gap. Within our framework, Background serves the purpose of stating the research area while providing historical, theoretical, or empirical context. Gap, however, concentrates on presenting prior studies and emphasizing their limitations, thereby justifying the study’s necessity and persuading the reader of its relevance and contribution to the field. Subsequently, we redefined the move category Product as Result to enhance its clarity in function. Finally, we restated the communicative function of Conclusion and introduced two new move categories: Implication and Contribution. The existing Conclusion category in H-2000, given its current scope, falls short of encompassing more specific functions when used as a label. Therefore, we introduced these two categories to amplify the granularity of move classification. These new categories cover sentences that serve the purposes of “drawing inferences that have not been explicitly stated in the abstracts” and “stating the theoretical and practical value of this paper.” This enrichment enables a more accurate understanding and analysis of abstracts’ moves and organizations. 

\begin{table*}[htbp]
    \centering
    \resizebox{\linewidth}{!}{
    \begin{tabular}{|l|l|}
    \hline
        \textbf{Move} & \textbf{Function}\\
    \hline
        Background& States the research area and provides any historical, theoretical, or empirical related information.\\
    \hline
        Gap& Establishes a niche: indicates a gap, adds to what is known, presents positive justification \cite{Swales:2004}.\\ 
    \hline
        Purpose& Indicates purpose, thesis or hypothesis, outlines the intention behind the paper.\\ 
    \hline
        Method& Provides information on design, procedures, assumptions, approach, data, etc.\\ 
    \hline
        Result& States main findings or results or what was accomplished.\\ 
    \hline
        Conclusion& Summarizes the results or extends results beyond scope of paper.\\
    \hline
        Implication& Draws inferences which has not been explicitly stated.\\
    \hline
        Contribution& Points out the theoretical and practical value.\\
    \hline
    \end{tabular}
    }
    \caption{Enriched move classification}
    \label{tab:table3}
\end{table*}


\section{Construction of the Corpus}
\subsection{Data Selection and Preprocessing}
\subsubsection{Data Selection}
Our corpus comprises abstracts carefully sampled from leading journals and internationally recognized conferences spanning the academic disciplines of AI and Engineering, in which the AI discipline includes NLP and Computer Vision(CV) domains; Engineering includes Communication Engineering (CE) and Mechanical Engineering (ME).
All of which have been selected with the invaluable guidance of experienced professors. The sources (Table~\ref{tab:table4}), are distinguished by their exceptional Journal Impact Factors and Q1 rankings in the Journal Citations Reports (JCR), unequivocally affirming their preeminent global standing.

Our choice of disciplines was guided by careful consideration. The critical status of Artificial Intelligence and Engineering in applied sciences informed our choice to engage in these fields. As researchers with a background in Natural Language Processing, our strategy was to start with the areas where we are most familiar with. We also identified and selected Computer Vision (CV) in AI, as well as Communication and Mechanical Engineering, as fundamental branches of Engineering to serve as data sources for our corpus construction. We will expand our corpus to cover more domains in the future.


\begin{table*}[htbp]
    \centering
    \resizebox{\linewidth}{!}{
    \begin{tabular}{|l|l|}
    \hline
      \textbf{Discipline} & \textbf{Journal/Conference} \\
   \hline
       Artificial Intelligence & the Annual Meeting of the Association for Computational Linguistics (ACL) \\
       Artificial Intelligence & Technical Track on CV on the AAAI Conference on Artificial Intelligence (AAAI) \\
       Mechanical Engineering & Journal of Mechanical Design \\
       Mechanical Engineering & International Journal of Heat and Mass Transfer \\
       Communication Engineering & IEEE Journal on Selected Areas in Communications \\
    \hline
    \end{tabular}
    }
     \caption{Selected journals and conferences for annotation}
    \label{tab:table4}
\end{table*}

\subsubsection{Data Preprocessing}
The ACL Anthology\footnote{https://aclanthology.org/} is the key resource that archives up-to-date conference papers in the CL and NLP domain. According to ACL Anthology's policy, all the materials are open-access and available for research purposes.
The website provides the metadata file (i.e., \textit{bib}) that can be downloaded directly. Considering the main conferences have the largest number of long papers, the length of abstracts in long papers typically ranges from 150 to 250 words, and the formal academic style is also the most standardized, we decided to preprocess the metadata file to extract abstracts from long papers in ACL main conferences held from 2020 to 2022. For the CV domain, abstracts in the CV track proceedings of AAAI 2022 are extracted. 

For journals, we first searched the name of each journal with the Web of Science (WOS) platform, and then the basic information of the articles in the journal including title, author, and abstract would be retrieved. After exporting and saving the retrieved results, we can collect all the abstracts data.

We performed sentence preprocessing on the abstract contents by dividing them into one sentence per line with the punctuation marks at the end of the sentences (such as periods) to facilitate the annotation later.

\subsection{Annotation Process}
\subsubsection{Annotation Guidelines}
The annotations were performed with the open-source and crowdsourcing text annotation tool doccano\footnote{https://github.com/doccano/doccano}. Based on the pilot study mentioned before, we defined 8 basic labels of move structures and their abbreviation for the annotation as shown in Table~\ref{tab:table5}. Besides, the labels are also visually distinguishable through the use of distinct colors in the annotation platform.

\begin{table}[htbp]
    \centering
    \begin{tabular}{|l|c|}
    \hline
        \textbf{Label} & \textbf{Abbreviation}\\
    \hline
        Background & BAC \\
        Gap & GAP \\
        Purpose & PUR \\
        Method & MTD \\
        Result & RST \\
        Conclusion & CLN \\
        Implication & IMP \\
        Contribution & CTN \\
    \hline
    \end{tabular}
    \caption{Annotation labels and their abbreviation}
    \label{tab:table5}
\end{table}

Principally, each complete sentence is assigned a single label, and annotators are asked to annotate the most suitable label for each sentence. Nevertheless, in case a sentence appears lengthy and contains two or more subsentences with distinct move structures, it is crucial to accurately label all the moves within that sentence. Here is an example:
\setlength{\parindent}{20pt}

[\textbf{BAC}]While neural networks with attention mechanisms have achieved superior performance on many natural language processing tasks, [\textbf{GAP}]it remains unclear to which extent learned attention resembles human visual attention.

\setlength{\parindent}{0pt}
\subsubsection{Manual Annotation}
To ensure high-quality annotation, our annotator team comprises four senior teachers and two graduate students specializing in linguistics and English language teaching. It is worth noting that, from the beginning of the annotation and construction of the corpus, the annotators held weekly discussions to collaboratively address challenges and difficulties that arose during the annotation process. This was especially crucial when it came to assigning appropriate labels to those controversial sentences. Through these discussions, the annotators gained a deeper understanding of the guidelines and the expressive characteristics of each abstract. The discussions foster a shared understanding and knowledge exchange within the expert annotators, enhance their comprehension of moves and professional skills in annotation and collaborative work, and allow for the documentation and learning from errors, leading to greater consistency and accuracy.
In this way, the conventional Inter-Annotator Agreement (IAA) testing was not employed during the annotation.

\begin{figure*}[htbp]
    \centering
    \includegraphics[width=\linewidth]{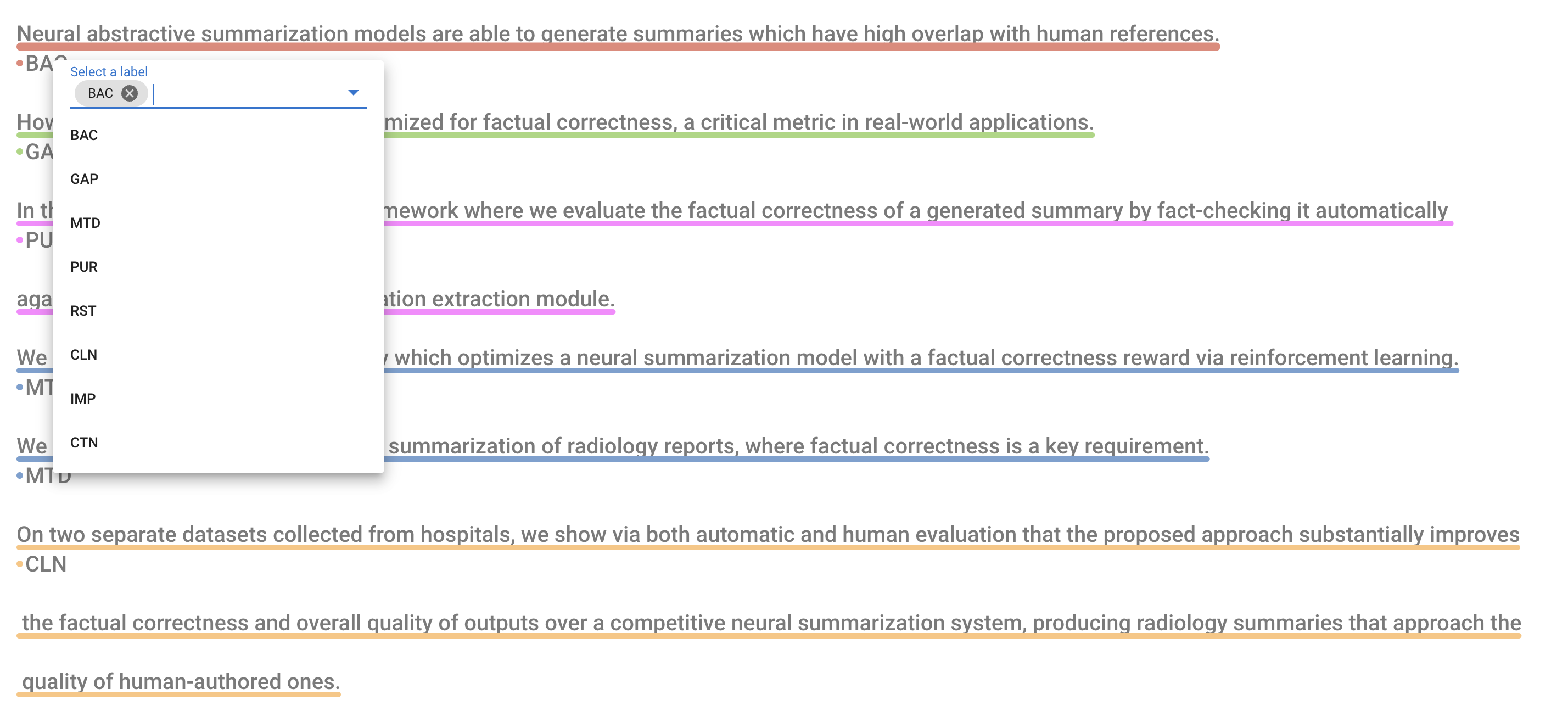}
    \caption{Screenshot of the doccano annotation platform}
    \label{fig:figure1}
\end{figure*}

Using the annotation platform (Figure~\ref{fig:figure1}), annotators can easily complete the annotation process by selecting an entire sentence and then choosing a pre-defined label. Once annotated, the label and a horizontal line matching the label’s color will appear below the sentence. If any modifications are necessary, annotators can simply click on the label beneath the sentence and choose a new one. The ID of abstracts, abstract texts and corresponding annotations can be exported and saved together in JSON format, as shown in the following example.

\texttt{\{
  "id": 20,\\
  "data": "Words can have multiple senses. Compositional distributional models of meaning have been argued to deal well with finer shades of meaning variation known as polysemy, but are not so well equipped to handle word senses that are etymologically unrelated, or homonymy.",\\
  "label": [[0, 31, "BAC"], [32, 265, "GAP"]]
\}}

\subsubsection{Automatic Annotation}
Despite the fact that manual annotations ensure high-quality, gold-standard data, they appear to be labor-intensive and time-consuming. To expedite the annotation process and scale up the corpus, we propose a BERT-based\cite{devlin2019bert} automatic annotation model with saliency attribution trained on the manual annotations. 

The model treats annotation as a multi-label recognition and classification problem. It learns semantic features from the annotated data, predicts move categories for unlabeled abstracts, and ultimately, assigns each sentence with a potential move label. 

Under certain circumstances, some words in the sentences may have direct impacts on predicting move labels. For instance, the appearance of the word “result(s)” may result in categorizing the sentence as either Result (RST) or Conclusion (CLN). Our proposed BERT model leverages saliency attribution to enhance move structure recognition\cite{lin2023move}. 

\begin{figure}[htbp]
    \centering
    \includegraphics[width=1\linewidth]{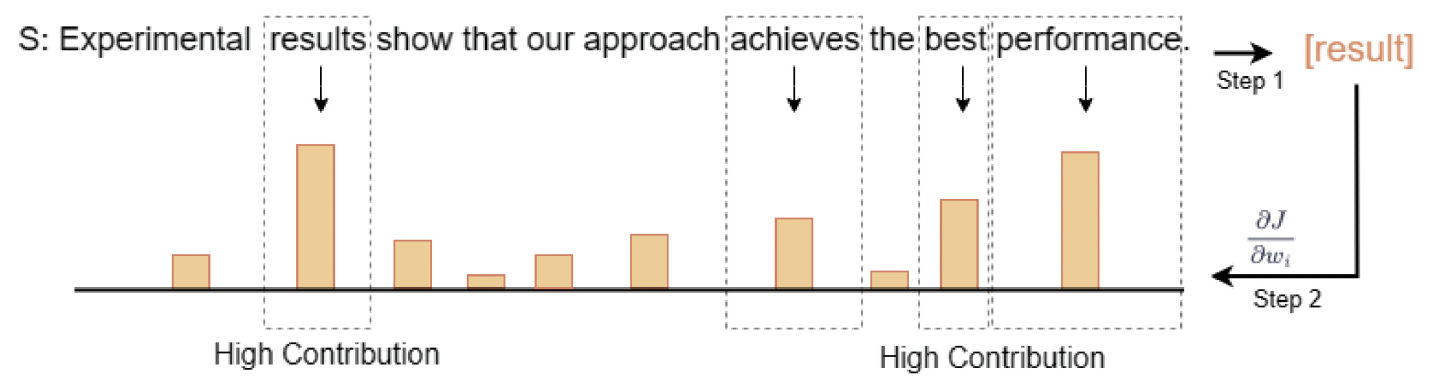}
    \caption{An illustration of move saliency attribution}
    \label{fig:figure2}
\end{figure}

To address such issues, our model integrates saliency attribution to improve the accuracy of move structure recognition. Initially, each sentence in the abstract is assigned a move label concerning the sentence’s overall meaning. Subsequently, each word within the sentences is regarded as a feature, and its contribution (saliency value) to a particular label is calculated. These saliency values ultimately serve as supporting evidence for enhancing move structure recognition. To sum up, the input of the BERT model contains four embedding vectors: namely Token Embeddings, Segment Embeddings, Position Embeddings, and Saliency Embeddings. An example of move saliency attribution is given in Figure~\ref{fig:figure2}. In this figure, a saliency value is assigned to each word to measure its impact on the overall semantic meaning of a move.

Once the automatic annotations were uploaded to the platform, annotators conscientiously reviewed them and corrected any errors present in the annotations. During our annotation practice, we found that there are some common trends in automatic annotation errors. Consider, for example, the frequent misidentification between PUR and MTD moves in model annotations. This occurs under two primary circumstances. One significant source of error is found in instances where abstracts begin directly with PUR moves without including any preceding background information (i.e., BAC and GAP), leading to an incorrect classification of these as MTD. 

Another prevalent misclassification is observed when authors of RAs commence their methodology descriptions with the phrase "we propose." In such cases, models tend to erroneously categorize these instances as PUR instead of MTD. 


Integrating feedback from annotators, the model was iteratively refined to enhance its performance, enabling the rapid expansion of the corpus while maintaining annotation quality.


\section{Corpus Statistics}
\subsection{Distribution of Move Types}
Under our framework of move analysis and implementing the aforementioned annotation procedure, all moves within the corpus have been identified and assigned to one of the eight predefined move types. We have annotated a total of 2,670 abstracts in the field of AI (1,340 in NLP and 1,330 in CV fields) and 2,003 abstracts in the Engineering field (1,003 in ME and 1,000 in CE).

Table~\ref{tab:table6} presents an overview of the 33,988 annotated move instances within the abstracts.

\begin{table}[htbp]
   \centering
   \begin{tabular}{|l|c|c|}
   \hline
      \textbf{Label} & \textbf{Frequency} & \textbf{\%}\\
   \hline
       BAC & 6,466& 19.02 \\
       GAP & 3,272& 9.63 \\
       PUR & 4,874& 14.34 \\
       MTD & \textbf{11,526}& \textbf{33.91} \\
       RST & 3,732& 10.98 \\
       CLN & 3,006& 8.84 \\
       IMP & 282& 0.83 \\
       CTN & 830 & 2.44 \\
    \hline
       \textbf{Total} & 33,988& 100 \\
   \hline
   \end{tabular}
   \caption{Frequency and distribution of moves identified in our corpus}
   \label{tab:table6}
\end{table}

The distribution of move structures provides compelling evidence of notable variations in the prevalence of move types within the corpus, characterized by pronounced disparities in both the frequency and proportion of these moves. Of particular significance is the substantial distinction between the move \textbf{MTD}, which holds the highest proportion at 30.68\% and significantly surpasses all other categories, indicating that most of the abstracts from various fields all focus on describing the methods and approaches, this distribution is also consistent with intuitive judgment, since methods are indeed one of the most attractive parts of abstracts. The least frequently occurring move \textbf{IMP}, which accounts for even less than 1\% of the total instances.



\subsection{Occurrence of Move Types}
It is pivotal to emphasize that move types are not equally well represented in the abstracts. Hence, we provide statistical examinations regarding the occurrence of these moves within the abstracts. To elaborate, when a specific move is identified in an abstract, regardless of its frequency, we consider it as \textbf{occurring} in that abstract. In addition, given the similar structural organizations of abstracts in CE and ME, we have combined them into the category of “engineering” and then proceeded to examine the occurrence of moves within the abstracts in these fields in comparison to the occurrence of moves in abstracts within the realm of AI. The moves’ occurrence in these two disciplines is presented in Table~\ref{tab:table7}.

\begin{table}[htbp]
    \centering    \begin{tabular}{|l|c|c|c|c|}
    \hline
 & \multicolumn{2}{c|}{\textbf{AI}}& \multicolumn{2}{c|}{\textbf{Engineering}}\\
    \hline
        \textbf{Move} & \textbf{\#} & \textbf{\%} & \textbf{\#} &\textbf{\%} \\
    \hline
       BAC & 2,003& 75.02  & 1,528&76.29  \\
       GAP & 1,518& 56.85  & 891 &44.48 \\
       PUR & \textbf{2,333}& \textbf{87.38}  & \textbf{1,901}& \textbf{94.91} \\
       MTD & 2,245& 84.08  & 1,873&93.51  \\
       RST & 1,540& 57.68  & 953 &47.58  \\
       CLN & 1,192& 44.64  & 1,079&53.87  \\
       IMP & 112 & 4.19  & 159 &7.94  \\
       CTN & 544 & 20.37  & 215 &10.73  \\
    \hline
    \end{tabular}
    \caption{Occurrence and distribution of each move type identified across the two fields in our corpus}
    \label{tab:table7}
\end{table}

Table~\ref{tab:table7} reveals some commonalities and differences in the occurrence of the moves across these two disciplines. For example, in both of the two fields, the move PUR occurs the most, and IMP occurs the least, PUR, MTD and BAC are top three moves ranking the occurrence. On the other hand, the proportion of PUR and MTD in the engineering field is far greater than those in the AI field, especially PUR. This discrepancy further emphasizes the potential divergence in move structures within abstracts across various disciplines, underscoring the necessity of investigating move structures in a variety of disciplines. This observation also indicates that moves themselves might differ in that some are obligatory in abstracts, some are optional but commonly present, while others seem to be optional and rare.


\begin{table}[htbp]
    \centering
    \resizebox{\linewidth}{!}{
    \begin{tabular}{|l|c|c|}
    \hline
         &  \textbf{AI} & \textbf{Engineering} \\
    \hline
    \textbf{\#Sent.}& 17,391& 16,597\\
         
    \hline
         \textbf{Average \#Sent.}&  6.51& 8.29\\
    \hline
         \textbf{\#Words}& 381,734& 406,244\\
         
    \hline
         \textbf{Average \#Words}&  142.97& 202.82\\
    \hline
         \textbf{Average \#Move types}&  4.38& 4.29\\
    \hline
    \end{tabular}
    }
    \caption{The average number of sentences, words, and move types in each abstract within the two fields}
    \label{tab:table8}
\end{table}

Table~\ref{tab:table8} provides a statistical overview of the sentence counts and occurrence of move types in each abstract from these two distinct disciplines. Abstracts’ sentence counts in the field of NLP range from 2 to 15, with an average of 6.5 sentences. Furthermore, the number of move types’ occurrence unveils that the majority of these abstracts comprise between 3 to 6 moves, and surprisingly, none of the abstracts contains the complete set of 8 move types. Abstracts in the engineering discipline exhibit longer sentence counts on average, with 8.3 sentences, indicating their lengthy composition. The number of move types in abstracts from these two disciplines, however, appears to be similar. These observations bolster our previous discussion on the potential obligatoriness or optionality of move types within an abstract.




\section{Experiments and Analysis}
In this section, we conduct move structure recognition experiments based on the annotated corpus to testify the effectiveness of the proposed model. We compare the results with baseline BERT and BERT+Context, where BERT+Context incorporates position information and the preceding and following sentences as additional contextual features. The performance is evaluated by Precision (P), Recall (R), and (Micro) F1.

We also compare the identification performance of our model with that of ChatGPT by manual evaluation.
\subsection{Move Structure Recognition}
The dataset and the results are shown in Table~\ref{tab:table9} and Table~\ref{tab:table10} respectively.

\begin{table}[htbp]
    \centering
    \begin{tabular}{|l|c|}
    \hline
     \textbf{Data} & \textbf{\#Sentences}\\
     \hline
     Training set & 7,147 \\
     Test set  & 1,787 \\
    \hline
    \end{tabular}
    \caption{Dataset statistics}
    \label{tab:table9}
\end{table}

\begin{table}[htbp]
    \centering
    \begin{tabular}{|l|c|c|c|}
    \hline
        \textbf{Method} & \textbf{P (\%)} & \textbf{R (\%)} & \textbf{F1 (\%)} \\
    \hline
    BERT    & 74.06 & 79.58 & 76.72\\
    BERT+Context     & 74.55 & 81.23 & 77.60 \\
    Our     & 75.01 & 82.34 & \textbf{78.53}\\
    \hline
    \end{tabular}
    \caption{Results of move structure identification}
    \label{tab:table10}
\end{table}

\begin{figure*}[htbp]
    \centering
    \includegraphics [width=0.9\linewidth]{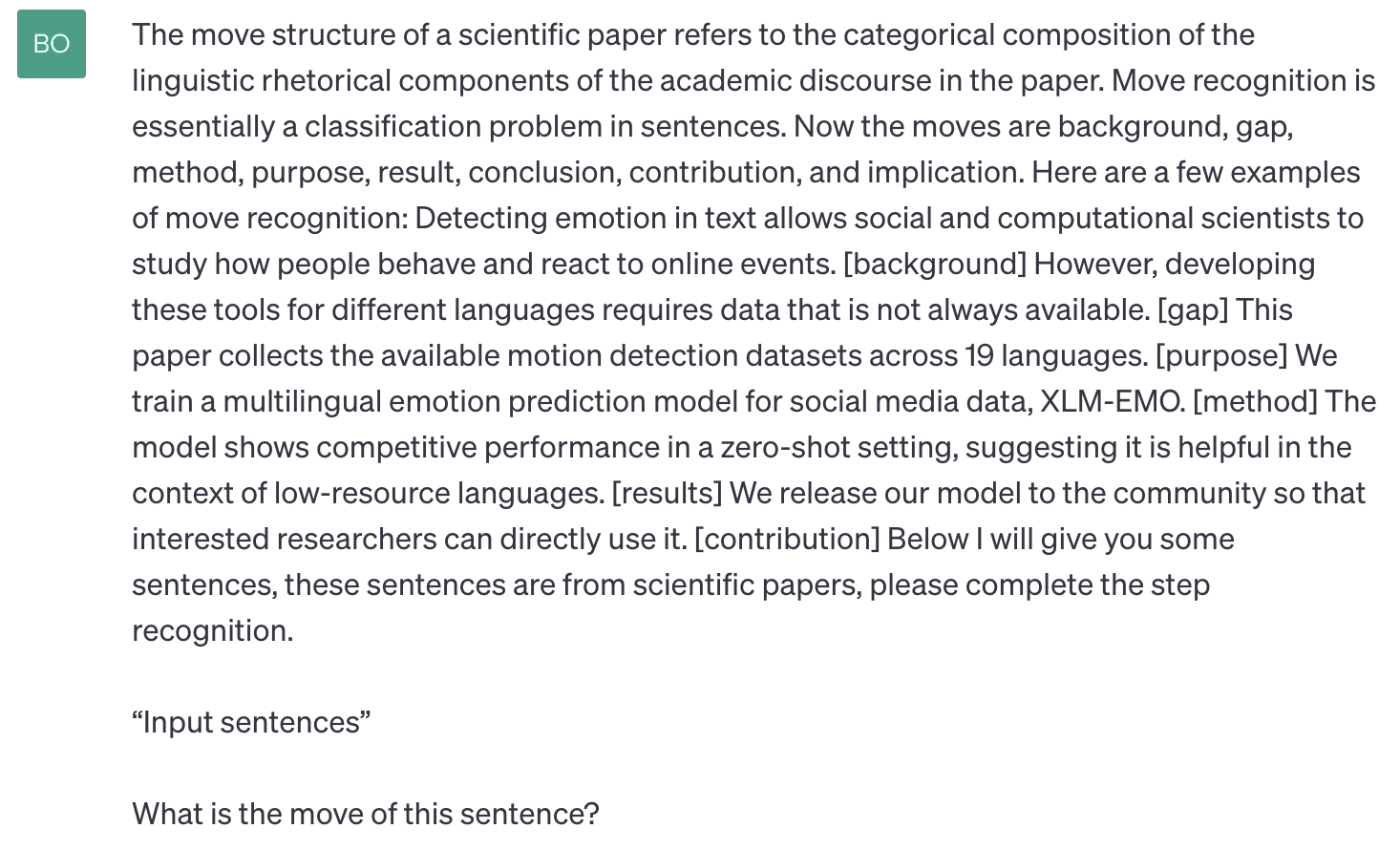}
    \caption{Instructions for ChatGPT}
    \label{fig:figure5}
\end{figure*}

\subsection{Comparison with ChatGPT}

The results indicate that our proposed method outperforms the baseline results, achieving the highest Micro F1 score (78.53\%), representing an improvement of 1.81\% over BERT and 0.93\% over BERT+Context, highlighting the model’s ability to not only account for the sequential relationship between contextual information and moves but also emphasize crucial structures in the sentences for accurate prediction and annotation of specific move types.

In this part, we compare the proposed model with ChatGPT. Our aim is to investigate whether Large language model (LLM) can achieve outstanding performance in move structure identification task and assess whether the proposed model could excel beyond the capabilities of LLM. 

We randomly selected 30 annotated abstracts from the AI and Engineering disciplines in our corpus, and employed both our proposed model and ChatGPT (gpt-3.5-turbo model) to identify the move structures. 

The process of conducting experiments through the gpt-3.5-turbo API is detailed as follows. Initially, a batch processing function capable of receiving multiple sentences as input and providing move structure recognition results for each sentence was defined. Following that, we formulated instructions for ChatGPT, as shown in Figure~\ref{fig:figure5}, to ensure the professionalism and accuracy of the experiment. Eventually, the move structure recognition results were obtained from ChatGPT.

Subsequent to the acquiring of both models’ results, we then asked three experts from each discipline, who were kept unaware of the identifications’ generating models, to manually evaluate the results with rating scores range from 0 to 100. 

Our model received an average score of 80, whereas ChatGPT obtained an average rating of 65. The results indicate that our proposed model excels at grasping the syntactic structures and context within the text, enabling more effective move structure recognition through the utilization of these structures in tandem with contextual information. Furthermore, our model has the potential for further refinement and optimization to align with specific situations and requirements, which may lead to more promising performance in specific tasks and domains than LLMs. 

\section{Conclusion}
In this study, we develop RAAMove, a multi-domain corpus specifically designed for the annotation of move structures within research article abstracts. Based on a modified theoretical foundation derived from Hyland's move classification, this paper delves into the scheme and construction process of the corpus. Notably, the construction process entails a combination of manual annotation and human-guided machine annotation. Through the development of clear annotation guidelines and the engagement of crowdsourced expert annotators, we construct high-quality data. Furthermore, we leverage the capabilities of a BERT-based model to automate data annotation, incorporating valuable feedback from human experts. This collective endeavor not only ensures the quality of the corpus but also expands its scale, making it a valuable resource for researchers.

Our preliminary experiments provide empirical evidence of the efficacy of the proposed corpus and its assoiated model. These findings emphasize the practical utility of RAAMove in various domains, including English language teaching, writing, and research article (RA) analysis. RAAMove is capable of facilitating tasks ranging from discourse and move analysis to move identification and intelligent-assisted correction of research articles. Moreover, we will release the corpus to the public in the near future, ensuring its accessibility to the academic community and beyond. 

In the future, we will continue to extend the annotation of move structures to abstracts across diverse domains, future enriching the depth of our linguistic analysis. In addition, we are considering the prospect of move annotation in other sections of research articles, such as the Introduction and Method sections. We are dedicated to furthering the boundaries of move structure analysis and look forward to the coming new developments in this field.

\section{Acknowledgements}
This work was supported by National Social Science Fund of China (NSSFC, 23BYY166). The authors thank the anonymous reviewers for their constructive comments and suggestions.

\section{Bibliographical References}\label{sec:reference}

\bibliographystyle{lrec-coling2024-natbib}
\bibliography{lrec-coling2024}

\end{document}